# Implementing Evidential Reasoning in Expert Systems

John Yen
USC / Information Sciences Institute
4676 Admiralty Way
Marina del Rey, CA 90292

**Abstract**

The Dempster-Shafer theory has been extended recently for its application to expert systems. However, implementing the extended D-S reasoning model in rule-based systems greatly complicates the task of generating informative explanations. By implementing GERTIS, a prototype system for diagnosing rheumatoid arthritis, we show that two kinds of knowledge are essential for explanation generation: (1) taxonomic class relationships between hypotheses and (2) pointers to the rules that significantly contribute to belief in the hypothesis. As a result, the knowledge represented in GERTIS is richer and more complex than that of conventional rule-based systems. GERTIS not only demonstrates the feasibility of rule-based evidential-reasoning systems, but also suggests ways to generate better explanations, and to explicitly represent various useful relationships among hypotheses and rules.[1]

## 1. Introduction

The Dempster-Shafer (D-S) theory has gained much attention in AI community in recent years [Garvey, Lowrance, and Fischler, 1981] [Sart, 1984][Gordon and Shortliffe, 1985] and has been applied to several applications regarding integrating data from multiple sources [Lowrance, Garvey, and Start, 1986]. However, the theory has not been implemented for reasoning in expert systems due to its difficulty dealing with uncertain rules. More recently, several extenstions to the theory has been proposed to overcome this difficulty [Yen, 1986a][Liu, 1986].

Based on Yen's extended D-S theory, we have implemented a prototype expert system, named GERTIS (General Evidential Reasoning Tool for Intelligent Systems), that diagnoses rheumatoid arthritis. We chose unspecified polyarthritis as the area of our medical consultation system because the diagnoses form a disease hierarchy, which fits Dempster-Shafer based reasoning best. GERTIS uses the knowledge base of CADIAG-2, a medical expert system developed by Peter Adlassnig [Adlassnig, 1985a,b]. Through the use of CADIAG-2's knowledge base, relevant evidence and rules have been already identified for the area of arthritis. In order to suit the needs of our model, however, the rules of CADIAG-2 were modified and reorganized.

One of the major difficulties in implementing GERTIS concern generating informative explanations and combining dependent evidence. Two kinds of knowledge turn out to be essential for explanation generation: (1) taxonomic class relationships between hypotheses, and (2) pointers to the rules that significantly contribute to belief in the hypothesis. As a result, the knowledge represented in GERTIS is richer and more complex than that of conventional rule-based systems.

---

[1]This article is based on the author's Ph.D. thesis at the University of California, Berkeley, which was supported by National Science Foundation Grant DCR-8513139.

In the next section, we describe how knowledge is represented in GERTIS, with an emphasis on its novelty. The explanation capabilities of GERTIS are discussed in section three, but the knowledge required for generating explanations is really covered in section two. A more detailed description of the system and its features can be found in [Yen, 1986b]. The theoretical aspects of the evidential reasoning model employed in GERTIS have been discussed in [Yen, 1986a], and, hence, is not duplicated here.

## 2. Knowledge Representation in GERTIS

Knowledge in GERTIS is represented by frames, hypotheses, and rules. Frames and rules correspond to the nodes and the links of an inference network. A hypothesis is a subset of its frame. The hypothesis object is mainly used to describe the class relationships between hypotheses within a frame. A rule links a group of *antecedent frames* to a *consequent frame*.

These objects in the knowledge base have their exact counterparts in the extended D-S theory [Yen, 1986a]. The frame represents the frame of discernment; the rule captures the probabilistic multivalued mapping from antecedent frame to the consequent frame; and the hypotheses are those focal elements that are of interest to experts. The following sections describe these objects in detail.

### 2.1. Frame Objects

A *frame object* represents a set of mutually exclusive and exhaustive possible values of a variable. For example, if a blood test result can be either positive or negative, the set {positive, negative} is a frame that contains all possible results of the blood test. Unlike frames in knowledge representation [Minsky, 1975], a frame object in GERTIS does not point to other frame objects. They are, instead, linked indirectly through rules. Most frames in GERTIS are binary frames (i.e., frames that contain exactly two elements). The only non-binary frame in GERTIS is the unspecified polyarthritis frame, which consists of all the system's final diagnoses.

The disease hierarchy of the unspecified polyarthritis frame, shown in Figure 2-1, contains nineteen terminal diagnoses (leaves), which are mutually exclusive and exhaustive. Here we assume that GERTIS is used for differential diagnosis after the patient has been diagnosed as having unspecified polyarthritis. In general, however, an "OK" node should be added to the hypothesis frame. Each node in the tree stands for a set of diagnoses. An internal node in the tree represents the union of its children nodes, and the leaf nodes represent single diagnoses. The black nodes represent hypotheses that are of interest to physicians, while the white nodes are not of particular interest to the physicians.

The computation of evidential reasoning involves many set operations among subsets of frames (i.e., focal elements in the D-S theory); hence, the internal representation of the subsets must support efficient implementation of these set operations. We choose to internally encode the focal element as bit vectors because both intersection operations and inclusion tests for sets can be efficiently implemented as bitwise logic operations on bit vectors. Within a frame, each bit position corresponds to an element of the frame. A value "1" in the position indicates that the element corresponding to the position is in the subset; a value "0" indicates the element is not in the subset.

A frame object stores five kinds of data: identification data, translation data, belief data, rule data, and structural data. *Identification data* include name and ID, a unique symbol used to identify and retrieve the frame object. *Translation data* are for the translation between symbolic names and bit-vector codes of the frame's subsets. *Belief data* include prior probability distribution, basic probability assignment (bpa), and basic certainty assignment (bca) of the frame. They are used to store and update the frame's belief function and are stored as association lists whose keys are focal elements' bit-vector codes. Since CADIAG-2's knowledge base does not contain experts' judgements about prior probabilities, we assume that the prior probability of polyarthritis frame is a uniform distribution. *Rule data* include a list of rules that the frame can trigger and a triggering history of the rules that update the frame's bpa. GERTIS is a forward chaining system; therefore, its frames contain an **f-rules** slot to store the frame's forward rules, rules whose antecedents refer the frame. The triggering history is used to retract the rule's conclusion when it becomes invalid due to a change of belief in evidence. See [Yen, 1986b] for a further discussion on rules retraction in GERTIS. *Structural data* provide access to the frame's hypothesis objects, which form a taxonomic class hierarchy similar to the disease hierarchy in Figure 2-1. Section four discusses how the hierarchical relationships between hypotheses assists GERTIS in generating better explanations.

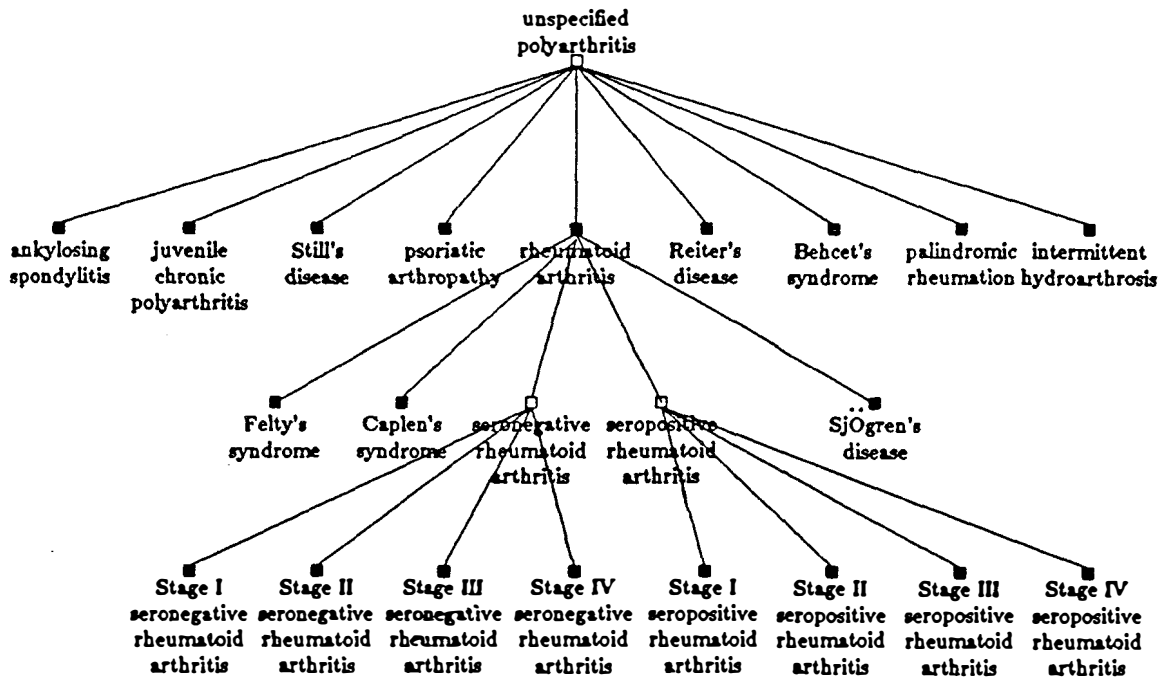

**Figure 2-1:** The Disease Hierarchy of Unspecified Polyarthritis

## 2.2. Rules

The conclusion of a GERTIS rule can be a set of hypotheses (or, equivalently, disjunctive hypotheses). In addition to this obvious difference, GERTIS rules differ from conventional rule-based systems in three major ways: (1) A GERTIS rule connects a group of antecedent frames to a consequent frame. (2) GERTIS removes taxonomic class relationships from rules, which represents inferential knowledge only. (3) In addition to numeric certainty, a GERTIS rule also contains a symbolic representation of its evidential strength. This section discusses these three differences in turn.

### 2.2.1. Antecedents and Consequents

Unlike the rules in MYCIN and PROSPECTOR, which link pieces of evidence directly to the hypothesis they bear upon, a GERTIS rule links a collection of *antecedent frames*, each of which contains a piece of evidence, to the *consequent frame*, which contains the hypothesis. The antecedent frames are referred in the rule premises in the form of "( antecedent-frame-name [ value ] )" with value defaults to "present." For example, the simple condition "(RE000007)" in rule R2 (Figure 3-1) states that "morning stiffness is present" because RE000007 is the name of the morning stiffness frame, which contains two elements (i.e., possible values): "present" and "absent." Making the distinction between the consequent frame and the hypotheses facilitates the seperation of local knowledge, knowledge about each individual hypothesis (e.g., class relationships), from global knowledge, knowledge about the whole frame of discernment (e.g., basic probability assignments).

The antecedent of a rule consists of an **IF** clause and an **EXCEPT** clause. While "IF" clauses describe the condition under which the rule is applicable, the "EXCEPT" condition describes situations that defeats the rule. As in CADIAG-2, antecedent conditions are combined using logic operators (e.g., AND, OR, NOT) and two special operators: the MIN and the MAX operators that describe elastic (fuzzy) constraints.

The consequent of a rule consists of a **THEN** clause and an **ELSE** clause. Each hypothesis (which corresponds to a focal element in D-S theory) in a "THEN" or "ELSE" clause is associated with the conditional probability of that hypothesis given the antecedent conditions. The *basic probability assignment* of the consequent frame is, thus, obtained from these conditional probabilities and the degree of belief in the antecedent.

### 2.2.2. Seperates Class Relationships from Inferential Knowledge

One of the advantages of GERTIS rule is that it seperates class relationships from inferential knowledge. The rules in expert systems have been criticized for representing contexts, control knowledge, and structural knowledge implicitly [Clancy, 1983][Swartout, 1983][Neches, Swartout, and Moore, 1984][Aikins, 1980]. Figure 2-2 shows a rule with implicit taxonomic class relationships between hypotheses in CADIAG-2. Rheumatoid arthritis appears in the antecedent of the rule because seronegative rheumatoid arthritis is one of its specializations (sub-diseases). GERTIS explicitly represents this kind of structural knowledge by coding sets as bit-vectors and by storing class relationships in the hypothesis objects. As a result, the rule in CADIAG-2 is replaced by a GERTIS rule R1, shown in Figure 2-3, whose antecedent does not contain rheumatoid arthritis. Moreover, the consequent of the CADIAG-2 rule is modified to consider other hypotheses in the frame. These hypotheses (the complement of "rheumatoid arthritis") were not inferred by the CADIAG-2 rule because they were implicitly excluded by the condition "rheumatoid arthritis." Thus, removing class relationships from a rule forces system builders to clearly characterize the association between the evidence and all the hypotheses in a frame.

```
IF rheumatoid arthritis, and
    latex agglutination test is negative
THEN seronegative rheumatoid arthritis
WITH strength of confirmation 1.0
```

**Figure 2-2:** A Rule in CADIAG-2 with Class Relationships

### 2.2.3. Roles

The *roles* of a rule serve two purposes. First, they specify the hypothesis that the rule has significant impact upon. In this paper, we shall call that hypothesis *acting hypothesis* of the rule. Second, they characterize the effect of the rule on its acting hypothesis using endorsement-like descriptions such as supportive, confirming, adversary, and disconfirming. Unlike the theory of endorsements [Cohen and Grinberg, 1983], however, GERTIS uses roles solely for generating explanations, not for reasoning. Furthermore, the endorsements are not meant to specify acting hypotheses as roles are.

Each GERTIS rule has a **t-role** and a **nil-role**, corresponding to the "THEN" clause and the

```
R1:
Consequent Frame: Unspecified Polyarthritis
IF: Latex agglutination test is negative
THEN: It is certain (1.0) that the patient has
      seronegative rheumatoid arthritis or
      NOT rheumatoid arthritis.
ELSE: no information
t-role: Support seronegative rheumatoid arthritis
nil-role: nil
```

**Figure 2-3:**   An Equivalent Rule in GERTIS without Class Relationships

"ELSE" clause respectively. Although symbolic roles have their counterparts in the consequents, the knowledge represented by roles, in general, is not redundant, i.e., it may not be deducible from the corresponding consequent clauses. For example, a t-role that can be deduced from THEN clause of R1 is "Confirming {seronegative rheumatoid arthritis or NOT rheumatoid arthritis}." However, the result of Latex result is never meant to be a piece of evidence against rheumatoid arthritis. The hypotheses in the complement of rheumatoid arthritis is mentioned in R1's consequent because they have no correlation with the test. Thus, it is possible for a patient to have these diseases whatever the test result might be. Consequently, only the "seronegative rheumatoid arthritis" is the acting hypothesis, although other hypotheses are possible given the test result. Furthermore, the numeric certainty (1.0) in R1's THEN clause becomes "Supportive", not "Confirming", in its t-role because the chances that the patient has seronegative rheumatoid arthritis given test result being negative is only relatively high, but not one, if we consider those hypotheses having no correlation with the test. In summary, obtaining roles of a rule from its corresponding consequent clause is greatly complicated by the large hypothesis space in evidential-reasoning systems. GERTIS, thus, relies on knowledge base builder to provide correct role information.

### 2.3. Hypothesis Objects

A *hypothesis object* is a subset of a frame that is of interest to domain experts. It is mainly used to store two kinds of knowledge for generating explanations: (1) hierarchical relationships to other hypotheses in the same frame, and (2) pointers to the rules that significantly contribute to belief in the hypothesis. In this section, we describe how the knowledge is stored in hypothesis objects.

The class relationships of hypotheses in a frame is explicitly represented through **superclass** and **subclasses** slots of hypotheses object. For example, the hypotheses of unspecified polyarthritis frame form a hypothesis taxonomy as illustrated in Figure 2-5. The class relationship in the taxonomy is used for generaing explanations (see section 3).

The knowledge about the rules that significantly contributes to the belief in a hypothesis is constructed from rules' role slots. For example, the rule R1 in Figure 2-3 supports seronegative rheumatoid arthritis when the rule condition is satisfied. Thus, R1 can make significant contribution to the belief in the diagnosis "the patient has seronegative rheumatoid arthritis." By recording R1 and its inferred basic probability assignment in the **triggered-b-rules** slot of seronegative rheumatoid arthritis when the rule is triggered, GERTIS can locate the rule efficiently to explain how the diagnosis is reached. Figure 2-4 depicts the hypothesis object of seronegative rheumatoid arthritis, assuming R1 was invoked with total belief (1.0) in its antecedent.

In Figure 2-4, the symbol "Rh" refers to the hypothesis object of rheumatoid arthritis. Similarly, the symbols Ne1, Ne2, Ne3, and Ne4 are symbolic names of Stage I, II, III, and IV seronegative rheumatoid

```
    name: Ne
    text: ``senegative rheumatoid arthritis''
    superclass: Rh
    b-rules: ( R1 )
    triggered-b-rules: ( ( R1 . ((522480 . 1.0)) ) )
    subclasses: (Ne1 Ne2 Ne3 Ne4)
```

Figure 2-4: A GERTIS Hypothesis Object

arthritis hypotheses respectively. The number "522480" is the internal bit-vector code of the subset "NOT {seronegative rheumatoid arthritis, Felty's syndrome, Caplen's syndrome, Sjogren's disease}."

## 2.4. Relationships Between GERTIS Objects

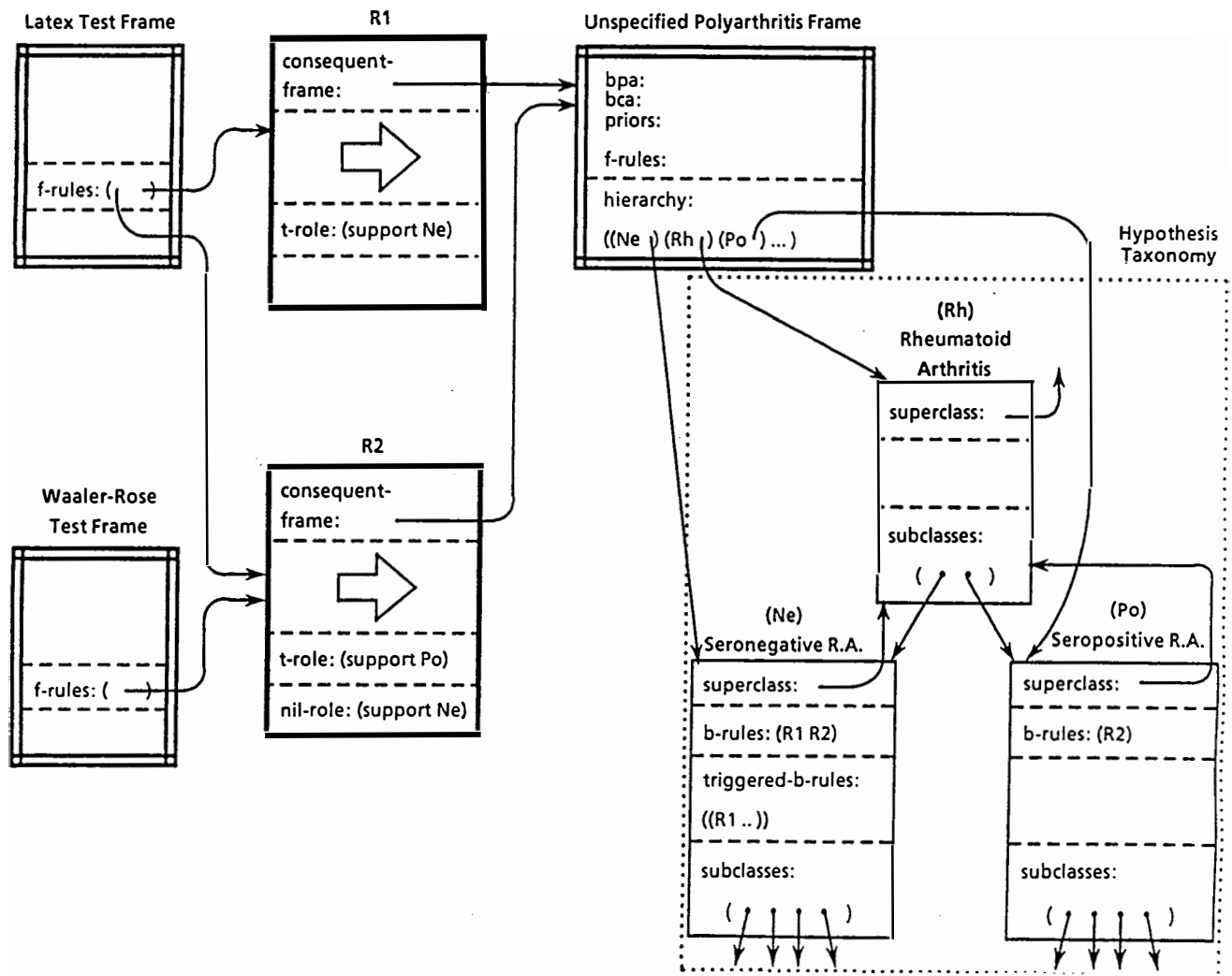

Figure 2-5: Relationships Between GERTIS Objects

The relationships between the three types of GERTIS objects are illustrated in Figure 2-5 and are summarized below.

- A rule connects several antecedent frames to a consequent frame. Frame objects keep a list of

forward rules in their **f-rules** slot. Hypothesis objects and their names are stored in the **hierarchy** slot of their frame object. Hypothesis objects are connected into a taxonomy through **superclass** and **subclasses** slots. Hypothesis objects maintain a list of backward rules in their **b-rules** slot and record those that get triggered in their **triggered-b-rules** slot.

## 3. Explanation Capabilities

Most expert systems generate explanations by tracing back through the rules that were triggered. This method in its simplest form generates too many explanations for GERTIS. For example, if GERTIS tries to generate explanations for the final diagnosis "The belief interval of seropositive rheumatoid arthritis is [0.7, 0.9]" by tracing back all the rules that affect the belief interval, the system may list too many rules for users to comprehend. Users would be even more confused because most rules listed would not seem to relate to the rheumatoid arthritis. Although all rules bearing on the unspecified polyarthritis frame affect the belief and plausibility of seropositive rheumatoid arthritis, only a small percentage of these rules make significant contribution to the diagnosis.

The questions are, then, (1) how to generate a more informative explanation, and (2) how to locate the rules that have major contributions to the belief interval of a given hypothesis. Our solutions are (1) using structural knowledge about hypothesis taxonomy to generate better explanations, and (2) using the **triggered-b-rules** slot in hypothesis object to locate the rules most relevant to the hypothesis.

GERTIS uses class relationships between hypotheses to generate intelligent explanations. For example, one of the reasons that the system infers seronegative rheumatoid arthritis is that the evidence collected strongly suggests rheumatoid arthritis, the superclass of seronegative rheumatoid arthritis. Thus, when GERTIS generates explanations for seronegative rheumatoid arthritis, it mentions rheumatoid arthritis and ask users if they want a further explanation of it. If the user types "yes," the same explanation procedure repeats for rheumatoid arthritis.

By recording the triggering history of a rule in its acting hypothesis, GERTIS is able to generate explanations for a diagnosis by tracing back only those rules whose roles refer to the diagnosis, i.e., those rules that have major contributioon to the diagnosis. Thus, the burden of locating right rules for explanation purpose is partially shifted from the system to knowledge engineers.

We illustrate the system's explanation capabilities by listing a brief consultation dialogue in Figure 3-1.

## 4. Conclusions

Implementing evidential reasoning in expert systems complicates the task of generating reasonable explanations. As a result, the knowledge base of GERTIS is augmented by structural knowledge about the hypothesis taxonomy and the symbolic representation of the "roles" that each rule plays. The roles of each rule are used to locate the rules most relevant to a diagnosis in question. The class relationships among hypotheses enable the system to use evidence of an abstract diagnosis as part of the explanation for its specializations.

GERTIS not only demonstrates that the extended D-S theory can be applied to expert systems, especially to those having multiple mutually exclusive hypotheses, but also suggests several domain-independent techniques for building rule-based evidential-reasoning systems.

```
    Command ? diagnose
Type the file name of patient symptoms (default file is Evidence) : E4
------------------------------------------------------------------
    Diagnostic Hypotheses                        Belief Intervals
------------------------------------------------------------------
    seronegative rheumatoid arthritis            [0.423,   1.000]

 Command ? why
    Number          Diagnosis
------------------------------------------------------------------
        0 =>    seronegative rheumatoid arthritis
Type the number of the diagnosis to be explaind : 0

The belief interval of seronegative rheumatoid arthritis is [0.423,   1.000]
is based on
(1) Rule1 and the observations that
    latex agglutination test is negative
                                        with degree of belief = 1.000,
and
(2) the belief interval of rheumatoid arthritis is [0.423,   1.000].

Do you want a further explanation of rheumatoid arthritis? (y or n)y

The belief interval of rheumatoid arthritis is [0.423,   1.000]
is based on
(1) Rule4 and the observations that
    (At lease  5 of the following symptoms are present:
        subcutaneous nodules over bone prominences, on extens or surfaces,
        or in intra-articular regions is present
                                        with degree of belief = 1.000,
        X-ray changes typical of rheumatoid arthritis is present
                                        with degree of belief = 0.700,
        positive rheumatoid factor is present
                                        with degree of belief = 1.000,
        poor mucin precipitate from synovial fluid is present
                                        with degree of belief = 1.000,
        characteristic histologic changes in synovial membrane is present
                                        with degree of belief = 1.000,
    )
    infers that the degree of belief in rheumatoid arthritis is  0.560,
and
(2) the belief interval of unspecified polyarthritis is [1.000,   1.000]

Do you want a further explanation of unspecified polyarthritis? (y or n)n
  Command ? quit
```

**Figure 3-1:** An Example of GERTIS Explanation Capabilities

## Acknowledgements

We would like to thank Professor Lotfi A. Zadeh for his continuous encouragement and support. We are mostly indebted to Dr. Peter Adlassnig at Department of Medical Computer Science, University of Vienna, and Dr. Gernot Kolarz at Ludwig Boltzmann Institute for Rheumatology and Focal Diseases, Baden, Austria for providing and translating the knowledge base of CADIAG-2. Many thanks also go to Professor Alice M. Agogino and Professor Richard Fateman at UC Berkeley, Dr. Robert Neches at USC/ISI, and Gerald Liu for their valuable comments on earlier drafts of the paper.

# References

1. K.-P. Adlassnig, "Present State of the Medical Expert System CADIAG-2," *Methods of Information in Medicine*, vol. 24 , pp. 13-20, 1985.

2. K.-P. Adlassnig, "CADIAG: Approaches to Computer-Assisted Medical Diagnosis," *Comput. Biol. Med.*, vol. 15, no. 5 , pp. 315-335 , 1985.

3. J. S. Aikins, "Prototypes and Production Rules: A Knowledge Representation for Computer Consultations," Report No. STAN-CS-80-814, Department of Computer Science, Stanford University, 1980.

4. W. Clancey, "The Epistemology of a Rule-Based Expert System: A Framework for Explanation," *Artificial Intelligence*, vol. 20, no. 3, pp. 215-251, 1983.

5. P. Cohen and M. Grinberg, "A Theory of Heuristic Reasoniong About Uncertainty," *The AI Magazine*, pp. 17-24, Summer 1983.

6. A. P. Dempster, "Upper and Lower Probabilities Induced By A Multivalued Mapping," *Annals of Mathematical Statistics*, vol. 38 , pp. 325-339, 1967.

7. T. D. Garvey, J. D. Lowrance, and M. A. Fischler, "An Inference Technique for Integrating Knowledge From Disparate Sources," *Proc. 7th International Conference on Artificial Intelligence*, pp. 319-325, 1981.

8. J. Gordon and E. H. Shortliffe, "A Method for Managing Evidential Reasoning in a Hierarchical Hypothesis Space," *Artificial Intelligence*, vol. 26 , pp. 323-357, 1985.

9. G. S. Liu, "Causal and Plausible Reasoning in Expert Systems," *Proc. 5th National Conference on Artificial Intelligence*, pp. 220-225, Philadelphia, PA, August, 1986.

10. J. D. Lowrance, T. Garvey, and T. M. Start, "A Framework for Evidential-Reasoning Systems," *Proc. 5th National Conference on Artificial Intelligence*, pp. 896-903, Philadelphia, PA, August, 1986.

11. M. Minsky, "A Framework for Representing Knowledge," in *The Psychology of Computer Vision*, ed. P. Winston, McGraw-Hill, New York, 1975.

12. R. Neches, W. Swartout, and J. Moore, "Enhanced Maintenance and Explanation of Expert Systems through Explicit Models of Their Development," *Transactions On Software Engineering*, November, 1985.

13. G. Shafer, *Mathematical Theory of Evidence,* Princeton University Press, Princeton, N.J., 1976.

14. T. M. Start, "Continuous Belief Functions for Evidential Reasoning," *Proc. 4th National Conference on Artificial Intelligence*, pp. 308-313, Austin, Texas, 1984.

15. W. Swartout, "XPLAIN: A System for Creating and Explaining Expert Consulting Systems," *Artificial Intelligence*, vol. 21, no. 3, pp. 285-325, September, 1983.

16. J. Yen, "A Reasoning Model Based on an Extended Dempster-Shafer Theory," *Proc. 5th National Conference on Artificial Intelligence*, pp. 125-131, Philadelphia, PA, August, 1986.

17. J. Yen, "Evidential Reasoning in Expert Systems," PhD thesis, Department of Electrical Engineering and Computer Science, University of California, Berkeley, September, 1986 .